\documentclass{cta-author}

{}
{}
{}

\begin{document}

\supertitle{Research Article}
\title{A Hybrid Tracking Control Strategy for an Unmanned Underwater Vehicle Aided with Bioinspired Neural Dynamics}

\author{\au{Zhe Xu$^1$} \au{Tao Yan$^1$} \au{Simon X. Yang$^1$} \au{S. Andrew Gadsden$^2$}}

\address{\add{1}{School of Engineering, University of Guelph, Guelph, Ontario, N1G 2W1, Canada}
\add{2}{Department of Mechanical Engineering, McMaster University, Hamilton, Ontario, L8S 4L8, Canada}
\email{syang@uoguelph.ca}}

\begin{abstract}
\looseness=-1 Tracking control has been a vital research topic in robotics. This paper presents a novel hybrid control strategy for an unmanned underwater vehicle (UUV) based on a bioinspired neural dynamics model. An enhanced backstepping kinematic control strategy is first developed to avoid sharp velocity jumps and provides smooth velocity commands relative to conventional methods. Then, a novel sliding mode control is proposed, which is capable of providing smooth and continuous torque commands free from chattering. In comparative studies, the proposed combined hybrid control strategy has ensured control signals smoothness, which is critical in real world applications, especially for an unmanned underwater vehicle that needs to operate in complex underwater environments.
\end{abstract}
\maketitle

\section{Introduction}\label{sec1}

The research on unmanned underwater vehicles (UUVs) has been ongoing for many years \cite{Zhang2020AdaptiveConstraints,Zhu2021AMap,Zhu2020AControl,Zhu2021Bio-inspiredCurrents,He2021RobustVehicles,Tijjani2021RobustExperiments,Gan2020ModelVehicles}. There are wide applications for UUV, such as ocean surveillance, fishing, and submarine construction surveys. Tracking control has always been a fundamental issue in UUV research, although there are many other challenging issues \cite{Zhu2015Multi-AUVPaper,Ni2017AVehicles}.

The studies on trajectory tracking control have been a major focus for autonomous vehicles. This research can be categorized primarily into four different methods: backstepping control \cite{Liang2018Three-DimensionalControl,Yang2012ARobots}, sliding mode control \cite{Qiao2020TrajectoryControl,Wu2021TrajectoryControl,Manzanilla2021Super-twistingVehicle,Zhu2022TrajectoryRepresentation}, Linearization control \cite{Chwa2010TrackingLinearization,Cheng2008ApplicationVehicle}, and neural networks and fuzzy logic control \cite{Sun2014,Wang2019CommandSpace,Zhang2021AdaptiveConstraints,Yu2020Guidance-Error-BasedDynamics}. Linearization control method \cite{Chwa2010TrackingLinearization,Kazemi2018FinitaryLinearization} is straightforward and easy to implement, however, this method faces difficulty when applied to UUVs. These difficulties arise due to the complex environments in which UUV's usually operate, which makes the hydrodynamic forces hard to obtain.

The backstepping control method \cite{Zhu2013TheVehicles,Peng2022ResearchStrategy,An2022AdaptiveQuantization} is one of the most commonly used control methods in robot control. The design of backstepping control aims to recursively stabilize a closed loop feedback system, which is relatively easy to design and proved using Lyapunov stability theory. In addition, backstepping control method has been adopted in UUV control design, however, the velocity control law is related to the tracking errors, therefore, the control signal will generate large initial velocity jumps if the initial tracking error is large. This large initial velocity change indicates that the UUV requires an impractical amount of torque to reach such velocities at initial stage.

Sliding mode control \cite{Zhang2018AAUV,Chen2021Finite-timeControl,Manzanilla2021Super-twistingVehicle} on UUVs has many advantages such as insensitivity to parameter variances and providing the system with extra robustness against disturbances. The downside of the sliding mode control design is that this control method has the chattering issue, and these high frequency changes in the control signal may damage the hardware in UUVs. The aforementioned velocity jump issue in backstepping control and chattering issue in sliding mode control could be dealt with using fuzzy logic control, which is another practical solution. However, the fuzzy logic control \cite{Yu2020Guidance-Error-BasedDynamics,Zhang2021AdaptiveConstraints} requires human knowledge and it is difficult to generate fuzzy rules that give satisfactory results. Neural networks \cite{Zong2021DecentralizedAUV,Fang2021NeuralAUVs} are another solution that has been implemented on UUV controls, which has been used to compute the complex nonlinear relations from ocean disturbances. However, this method requires online learning and training processes, which could be expensive and computationally complicated.

The bioinspired neural dynamics model was first proposed by Hodgkin and Huxley \cite{Hodgkin1952ANerve} based on a patch of membrane in a biological neural system using electrical circuit elements. Then, this bioinspired neural dynamics model was first used by Yang \cite{Yang2000AnPlanning} in robotics, which was later expanded into many other applications \cite{Yang2003Real-TimeApproach,Yang2001NeuralGeneration,Yang2019DisturbanceRobots}. In the following work from Zhu \cite{Zhu2013TheVehicles}, this bioinspired neural dynamics model was implemented on a kinematic controller for a UUV in path tracking problems. The entire control design was a hybrid control, which used a bioinspired kinematic controller to resolve the speed jump issue in the conventional backstepping controller and a sliding mode controller for the dynamic control of the UUV.

Inspired by the special characteristic of the bioinspired neural dynamics, this paper designs a bioinspired sliding mode controller that is capable of resolving the chattering issue, which is more practical in real-world applications. The chattering issue has existed since sliding mode control was developed, which can be observed from many applications. Although the chattering issue has been solved, such as using saturation or tanh function to replace the chattering term, there are still some drawbacks to the existing solutions, such as losing the robustness to the disturbances and the parameters have to be perfectly tuned to prevent chattering; in addition, considering the noises, the conventional backstepping and sliding mode control are sensitive to the noises, whereas the bioinspired based controller is capable of providing smooth control command, which is critical due to the limitations of the actuator.

Therefore, in this paper, a novel hybrid control method aided with the bioinspired neural dynamics model is proposed based on the preliminary work in \cite{Xu2021BacksteppingNeurodynamics}. This overall design contains a kinematic controller and a dynamic controller that are integrated with the bioinspired neural dynamics model. The proposed bioinspired kinematic controller is capable of avoiding the velocity jump. Then, to avoid chattering issue in conventional sliding mode control, the same bioinspired neural dynamics is combined with the sliding mode control to provide continuous and smooth dynamic control output that is free from chattering. In addition, the proposed control strategy is capable of providing smooth control commands under the system and measurement noises due to the filtering capability from the bioinspired neural dynamics.

This paper is organized as follows, Section II provides the kinematic and dynamic model of a UUV operating underwater. Then, Section III provides the bioinspired neural dynamics model that helps to develop the bioinspired backstepping kinematic controller and bioinspired sliding mode controller. The stability analysis of the proposed control strategy is provided in Section III as well. After that, Section IV gives the results, which are compared with multiple other methods to demonstrate its efficiency and effectiveness. Finally, the conclusion in Section V illustrates the developed concept and why the proposed control method is a better control strategy for a UUV. Some potential future improvements are also mentioned in the conclusion.

\section{Kinematics and Dynamics of a UUV}
The UUV's dynamic model is presented as \cite{Soylu2008AAllocation}
\begin{equation}
    M\dot V + C(V)V + D(V)V + g(\eta) = \tau \label{1}
\end{equation}
where $M$ is the inertial matrix, $C$ is the Coriolis and centrifugal forces, and $D$ is the hydrodynamics force. The UUV's gravity and buoyancy forces are represented by $g$. Figure 1 demonstrates the coordinate systems for UUV, the inertial frame of the UUV is represented by $\left\{ {X,Y,Z,O} \right\}$ and $\left\{ {{X_0},{Y_0},{Z_0},{O_0}} \right\}$ is the coordinate of the body fixed frame coordinate system. The position and orientation of the UUV in inertial frame is represented by vector $\eta  = {[\begin{array}{*{20}{c}}
x&y&z&\phi&\theta&\psi \end{array}]^T}$, where $x$, $y$, and $z$ are the position and $\phi$, $\theta$, and $\psi$ are the orientation in $X$, $Y$, and $Z$, respectively. The velocity vector of the UUV in body fixed frame is represented by $V = {[\begin{array}{*{20}{c}}u&v&w&p&q&r
\end{array}]^T}$, where the elements in $V$ are the surge, sway, heave, roll, pitch, and yaw motions of the UUV. The controlled torque is defined as $\tau$ that acts on the mass center of the UUV. 
\begin{figure}[h]
\centerline{\includegraphics[width=0.45\textwidth]{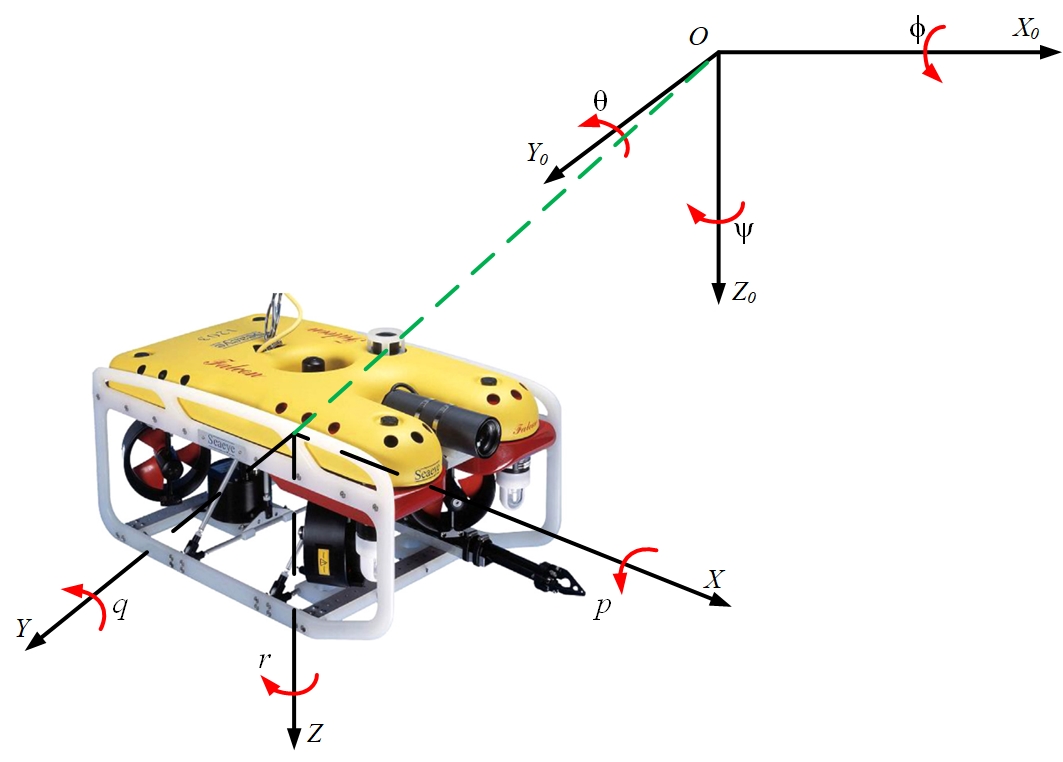}}
\caption{UUV coordinates in body fixed and inertial frame.\label{fig1}}
\end{figure}The kinematic of the UUV is given as
\begin{equation}
\label{eq2}
    \dot \eta  = J\left( \eta  \right)V
\end{equation}
where $J(\eta)$ is a transformation matrix between the body fixed frame and the inertial frame. This paper discusses the UUV operates in horizontal plane; therefore, the reduced dynamics of the UUV can be written in a decoupled form as \cite{Lapierre2008RobustAUV}
\begin{equation}
\label{eq1}
    {\tau _x} = \left( {m - {X_{\dot u}}} \right)\dot u + {X_u}u + {X_{uu}}u\left| u \right| \vspace{-1ex}
\end{equation}
\begin{equation}
   {\tau _y} = \left( {m - {Y_{\dot v}}} \right)\dot v + {Y_v}v + {Y_{vv}}v\left| v \right|
\end{equation}
\begin{equation}
   {\tau _N} = \left( {{I_z} - {N_{\dot r}}} \right)\dot r + {N_r}r + {N_{rr}}r\left| r \right|
\end{equation}
where $m$ represents the mass, and added mass effects of $u$ and $v$ are ${X_{\dot u}}$ and ${Y_{\dot v}}$, respectively. Parameter ${N_{\dot r}}$ is the added moment of inertia from $r$, and ${I_z}$ is the moment of inertia in $z$. Linear drag and in $u$, $v$, and $r$ are, respectively, $X_u$, $Y_v$, and $N_r$. The quadratic drags and in $u$, $v$, and $r$ are  $X_{uu}$, $Y_{vv}$, and $N_{rr}$, respectively. The quadratic drags and in $u$, $v$, and $r$ are  $X_{uu}$, $Y_{vv}$, and $N_{rr}$, respectively. Vector $\tau {\rm{ = }}{\left[ {\begin{array}{*{20}{c}}
{{\tau _x}}&{{\tau _y}}&{{\tau _N}}
\end{array}} \right]^{\rm{T}}}$ is the control force and moment that is applied on the center of mass on UUV in $X$,$Y$ and orientation of body fixed frame.      
\begin{figure*}[t!]
\centerline{\includegraphics[width=1\textwidth]{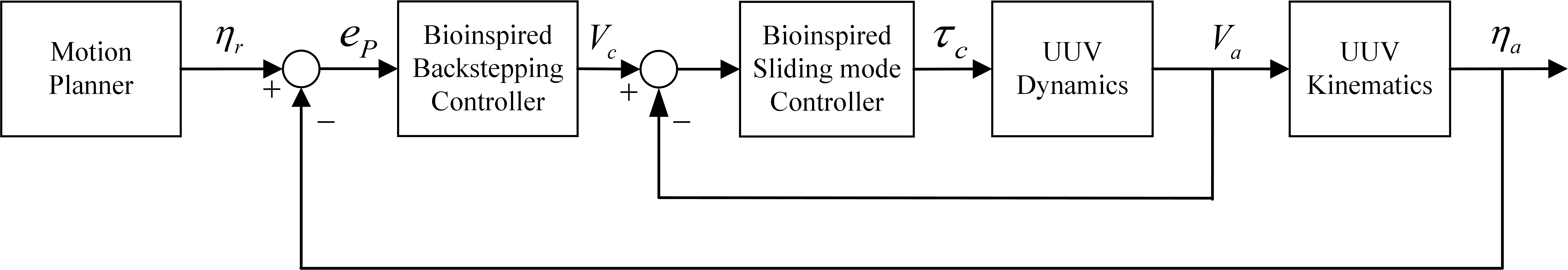}}
\caption{Block diagram of a UUV based on the bioinspired control strategy}\label{fig2}
\end{figure*}

\section{UUV Control Design Aided with Bioinspired Neural Dynamics}
This section develops a hybrid control strategy for a UUV. The block diagram in Figure \ref{fig2} demonstrates the designed control, which is a hybrid control strategy that consists two closed loops. The outer loop consists a bioinspired backstepping based kinematic controller that is capable of providing smooth velocity command and preventing sharp velocity jump from sudden change of initial tracking errors. Then, the output control command from the bioinspired backstepping controller is processed through the inner loop control, which is a sliding mode controller that is also combined with the bioinspired neural dynamics. The bioinspired sliding mode controller is capable of preventing the chattering issue that occurs in conventional sliding mode control, yet a certain amount of robustness to disturbances remains as the bioinspired neural dynamics acts like a low-pass filter.
\subsection{Bioinspired neural dynamics model}\label{neuro}
Hodgkin and Huxley \cite{Hodgkin1952ANerve} proposed a membrane model based on a path of membrane using electrical elements. This model was then developed by Grossberg \cite{Grossberg1988NonlinearArchitectures} that described a real-time adaptive behaviour of individuals. Then, it was first applied in robotics by Yang and Meng \cite{Yang2000AnPlanning}, which has been expanded into many other robotic applications. The bioinspired inspired neural dynamics model, which is called shunting model is written as
\begin{equation}
    \dot x_i =  - {A_i}{x_i} + ({B_i} - {x_i}){S_i}^ + - ({D_i} + {x_i}){S_i}^ -\label{6}
\end{equation}
where $x_i$ is the neural activity of $i$-th neuron, $A_i$ is the passive decay rate and $B_i$ and $D_i$ are, respectively, the upper bound and lower bound of the $i$-th neuron. Variables ${S_i}^ +$ and ${S_i}^ -$ are the $i$-th neurons excitatory and inhibitory inputs. Based on \eqref{6}, the shunting model that is applied in this paper is defined as
\begin{equation}
    {\dot L_i} =  - A_i{L_i} + (B_i - {L_i}){f}({e_i}) - (D_i + {L_i}){g }({e_i})\label{shut}
\end{equation}
where $e_i$ is the error between the reference state and the actual state of the UUV, which is treated as input of the bioinspired neural dynamics, ${f }({e_i})$ and ${g }({e_i})$ are defined as ${f }({e_i})=\text{max}(e_i,0)$ and ${g}({e_i})=\text{max}(-e_i,0)$, respectively. Variable $L_i$ is the output, which is guaranteed to be bounded within $({-D_i},{B_i})$ for any input of $e_i$. In addition, it is worth noting that the shunting model also has the filtering property that acts like a low-pass filter. Equation \eqref{shut} is applied twice in this paper to separately address the velocity jump and chattering issues in the backstepping control and the sliding mode control. Inspired by the special characteristic of the shunting model, the following subsections explain the design procedures in detail.
\begin{figure}[b]
\centerline{\includegraphics[width=0.3\textwidth]{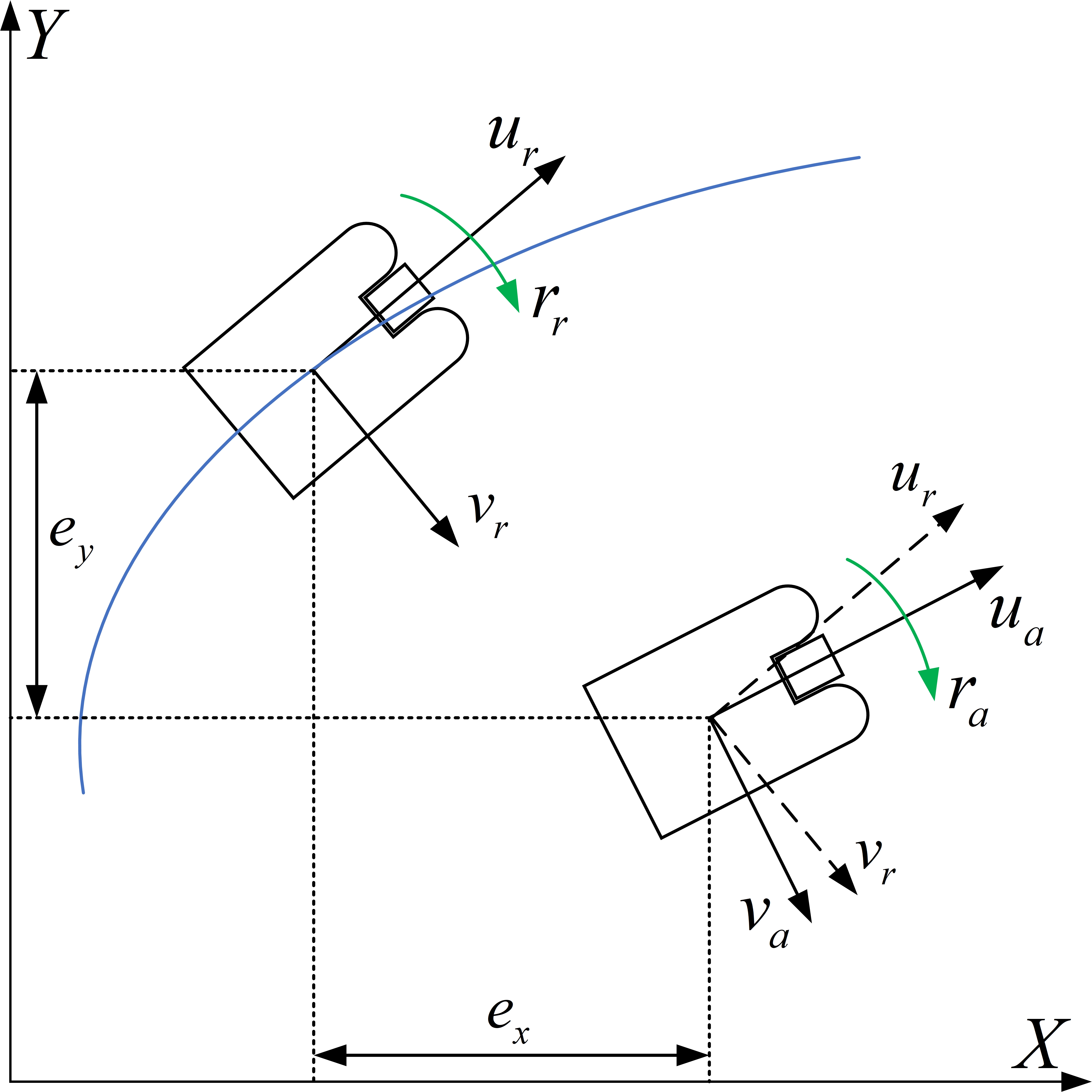}}
\caption{Coordinate conversion for UUV.\label{fig3}}
\end{figure}
\subsection{Bioinspired backstepping kinematic control}
The backstepping control design is straightforward and relatively easy to design. This paper discusses the motion of a UUV in horizontal plane, therefore, for a UUV operating horizontally, the reference posture of a UUV is defined as ${\eta _r} = {\left[ {\begin{array}{*{20}{c}}
{{x_r}}&{{y_r}}&{{\psi _r}}
\end{array}} \right]^T}$ in inertial frame. The desired reference velocity state in surge, sway, and yaw motions are defined as ${V_r} = {\left[ {\begin{array}{*{20}{c}}
{{u_r}}&{{v_r}}&{{r_r}} \end{array}} \right]^T}$. Based on \eqref{eq2}, the relations between the reference position state and the reference velocity state can be obtained through a transformation matrix $J(\eta)$, which is defined as
\begin{equation}
    \left[ {\begin{array}{*{20}{c}}
{{u_r}}\\
{{v_r}}\\
{{r_r}}
\end{array}} \right] = \left[ {\begin{array}{*{20}{c}}
{C {\psi _r}}&{S {\psi _r}}&0\\
{ - S {\psi _r}}&{C {\psi _r}}&0\\
0&0&1
\end{array}} \right]\left[ {\begin{array}{*{20}{c}}
{{{\dot x}_r}}\\
{{{\dot y}_r}}\\
{{{\dot r}_r}}
\end{array}} \right]\label{jac}
\end{equation}
where $S$ and $C$ represent $\sin$ and $\cos$ function, respectively. The objective of the kinematic controller is to generate velocity control commands, which forces the UUV to reach a certain velocity that makes the tracking error approach zero. The tracking error with respect to the body fixed frame is defined as ${e_P} = {\left[ {\begin{array}{*{20}{c}}
{{e_1}}&{{e_2}}&{{e_3}}
\end{array}} \right]^T}$, where $e_1$, $e_2$, and $e_3$ are the tracking error in surge, sway, and yaw motion with respect to the body fixed frame. The tracking error with respect to inertial frame is calculated by
\begin{equation}
    {e_P} = J{(\eta )^{{\rm{ - }}1}}{\left[ {\begin{array}{*{20}{c}}
{{e_x}}&{{e_y}}&{{e_\psi}}
\end{array}} \right]^T}\label{7}
\end{equation}
Figure \ref{fig3} shows the coordinates conversion of the tracking error between the fixed body frame and the inertial frame, where $e_x$,  $e_y$, and  $e_\psi$ are, respectively, the tracking error in $X$,$Y$, and the orientation in the inertial frame. Then, the convectional design of the backstepping based kinematic control for the UUV is given as 
\begin{equation}
\begin{array}{l}
{V_c} = {\left[ {\begin{array}{*{20}{c}}
{{u_c}}&{{v_c}}&{{r_c}}
\end{array}} \right]^T} = \\
\left[ {\begin{array}{*{20}{c}}
{{k_a}\left( { {e_x}C (\psi)  + {e_y}S (\psi) } \right) + \left( {{u_r}S ({e_\psi }) - {v_r}C ({e_\psi })} \right)}\\
{{k_a}\left( { - {e_x}S (\psi)  + {e_y}C (\psi) } \right) + \left( {{u_r}S ({e_\psi }) + {v_r}C ({e_\psi })} \right)}\\
{{r_d} + {k_b}{e_\psi }}
\end{array}} \right]
\end{array}\label{8}
\end{equation}
The velocity commands in the surge, sway, and yaw motions are $u_c$, $v_c$ and $r_c$, respectively. Parameters $k_a$ and $k_b$ are the control parameters for the backstepping based kinematic controller. Based on \eqref{8}, it can be observed that if there is a large tracking error at the initial stage, ${ {e_x}C (\psi)  + {e_y}S (\psi) }$, ${ - {e_x}S (\psi)  + {e_y}C (\psi) }$, and ${k_b}{e_\psi }$ will cause speed jumps in surge, sway, and yaw motions, respectively. Therefore, in order to resolve the speed jump issue, the bioinspired backstepping control is proposed, which the new virtual velocity command is defined as
\begin{equation}
\begin{array}{l}
{V_c} = {\left[ {\begin{array}{*{20}{c}}
{{u_c}}&{{v_c}}&{{r_c}}
\end{array}} \right]^T} = \\
\left[ {\begin{array}{*{20}{c}}
{{k_a}\left( {{L_1}C (\psi)  + {L_2}S (\psi) } \right) + \left( {{u_r}S ({e_\psi }) - {v_r}C ({e_\psi })} \right)}\\
{{k_a}\left( { - {L_1}S (\psi)  + {L_2}C (\psi) } \right) + \left( {{u_r}S ({e_\psi }) + {v_r}C ({e_\psi })} \right)}\\
{{r_r} + {k_b}{L_3}}
\end{array}} \right]
\end{array} \label{11}
\end{equation}
where $L_1$, $L_2$ and $L_3$ are the outputs from the bioinspired neural dynamics whereas $e_x$, $e_y$, and $e_\psi$ are the respected inputs.  Due to the special characteristics of the shunting model, by assuming $B_i=D_i$, the error term in the virtual control commands have the following inequalities
\begin{equation}
\begin{aligned}
    {k_a}({L_1}C(\psi ) + {L_2}S(\psi )) &< {k_a}\left( {{B_1} + {B_2}} \right)\\
    {k_a}(-{L_1}C(\psi ) + {L_2}S(\psi )) &< {k_a}\left( {{-B_1} + {B_2}} \right)\\
    {k_b}{L_3}&<{k_b}{B_3}
\end{aligned}
\end{equation}where $B_1$, $B_2$, and $B_3$ are positive constants. Thus, by carefully select the $u_r$, $\upsilon_r$, and $r_r$, the velocity virtual control inputs are bounded and not exceeding its dynamic constraints. In addition, the bioinspired backstepping control is also capable of providing smooth control input under noises due to its filtering capability, $A_i$ in the shunting model acts as the bandwidth of the low pass filter. 
To prove the stability of the bioinspired backstepping control, the Lyapunov candidate function is proposed as 
\begin{equation}
    {V_p} = \frac{1}{2}\left( {e_x^2 + e_y^2 + e_\psi ^2} \right) + \frac{k_a}{{2B_1}}L_1^2 +\frac{k_a}{{2B_2}}L_2^2+ \frac{{{k_b}}}{{2B_3}}L_3^2
\end{equation}
then, the derivative of $V_p$ is found to be
\begin{equation}
    {{\dot V}_p} = {e_x}{{\dot e}_x} + {e_y}{{\dot e}_y} +{e_\psi}{{\dot e}_\psi}+ \frac{k_a}{{B_1}}{L_1}{{\dot L}_1} + \frac{k_a}{{B_2}}{L_2}{{\dot L}_2} + \frac{k_b}{{B_3}}{L_3}{{\dot L}_3} \label{22}
\end{equation}
The proposed bioinspired kinematic control strategy is capable of generating smooth velocity commands and avoiding sharp velocity jumps if an initial tracking error occurs, and its stability will be proven in Subsection \ref{stable}.
\subsection{Bioinspired sliding mode dynamic control}
This subsection develops a sliding mode controller that is aided with the bioinspired neural dynamics aforementioned from Subsection \ref{neuro}. The generated virtual velocity commands from bioinspired backstepping control are treated as the reference state for the dynamic control of the UUV. The control command $\tau$ is then generated from bioinspired sliding model controller, which is used to drive the dynamic model of UUV to reach its desired state.

First, the velocity error is defined $e=V_c-V_a$, where $V_a$ is the actual velocity that is observed with respect to the body fixed frame. Defining $\Gamma$ as a positive constant, then the sliding surface is proposed as 
\begin{equation}
   S = \dot e + 2\Gamma e + {\Gamma ^2}\int e \label{12}
\end{equation}
By taking the derivative of \eqref{12}, which gives
\begin{equation}
      \dot S = \ddot e + 2\Gamma \dot e + {\Gamma ^2}e = \ddot e + 2\Gamma ({{\dot V}_c} - {\dot V}_a) + {\Gamma ^2}e \label{13}
\end{equation}
Based on \eqref{13}, if the system operates on the sliding surface, then, 
\begin{equation}
    \dot S = \ddot e + 2\Gamma ({{\dot V}_c} - {\dot V}_a) + {\Gamma ^2}e = 0 \label{14}
\end{equation}
From \eqref{1} into \eqref{14}, it is calculated as
\begin{equation}
    \ddot e + 2\Gamma ({{\dot V}_c} - {M^{ - 1}}(\tau  - C{V_a} - D{V_a} - g)) + {\Gamma ^2}e = 0 \label{15}
\end{equation}
Therefore, the equivalent control law is found that
\begin{equation}
    {\tau _{eq}} = M({{\dot V}_c} + \frac{{\ddot e}}{{2\Gamma }} + \frac{\Gamma }{2}e) + C{V_a} + D{V_a} +  g \label{16}
\end{equation}
A feedback control input of acceleration is used to in \eqref{16} due to the difficulty of computing $\ddot e$, which is defined as
\begin{equation}
    \ddot e = {k_s}\dot e
\end{equation}
Therefore, the conventional sliding mode control law is defined as 
\begin{equation}
    \tau  = {\tau _{eq}} + k{\mathop{\rm sgn}} (S) \label{chat}
\end{equation}
It is observed from \eqref{chat} that the chattering in control signal is caused by sign function. One commonly used control method that is capable of eliminating the chattering issue in conventional sliding mode control is by replacing the sign function with saturation function. The sliding mode control with saturation function can be defined as
\begin{equation}
\begin{array}{c}
\tau  = {\tau _{eq}} + {\rm{Sat}}(S)\\
{\rm{Sat}}(S) = \left\{ {\begin{array}{*{20}{c}}
{{B_4}}&{{\rm{if}}}&{{k_s}S > {B_4}}\\
{{k_s}S}&{{\rm{if}}}&{ - {D_4} \le {k_s}S \le {B_4}}\\
{ - {D_4}}&{{\rm{if}}}&{{k_s}S <  - {D_4}}\label{sat}
\end{array}} \right.
\end{array}
\end{equation}
where $k_s$ is a positive constant that defines the changing rate of the control output, $B_4$ and $-D_4$ are the upper bound and lower bound of the saturation function, respectively. Although the saturation function works in some situations if the control parameters are relatively small, UUVs usually operate in complicated environments, which sometimes require larger control parameters. Therefore, the same bioinspired neural dynamics that applied to the conventional backstepping control in \eqref{11} is combined with sliding mode control again. The bioinspired sliding mode control is defined as
\begin{equation}
    \tau  = {\tau _{eq}} + {L_4}\label{18}
\end{equation}
where $L_4$ is the output of the shunting model that is defined in \eqref{6}. The Lyapunov candidate function for the bioinspired sliding mode controller is proposed as 
\begin{equation}
    V_z=\frac{1}{2}{S^T}S + \frac{1}{{2{B_4}}}{L_4}^T{L_4} \label{21}
\end{equation}
Then, the derivative of \eqref{21} is calculated as
\begin{equation}
    \dot V_z={S^T}\dot S + \frac{1}{{{B_4}}}{{\dot L}_4}^T{L_4}\label{25}
\end{equation}
The proposed sliding mode dynamic control input solved the chattering issue that occurs in the conventional control method, yet gives extra robustness to the controller since there are more parameters that can be tuned for the controller to reach satisfactory results. 

\subsection{Stability analysis}\label{stable}
The proposed control strategy contains two closed loops, which are bioinspired backstepping kinematic control in outer loop and bioinspired sliding mode control in inner loop. These two kinematic and dynamic controls are separately proven to be asymptotically stable using Lyapunov stability analysis. Then, an overall Lyapunov candidate function is provided to demonstrate that the proposed control strategy reaches global asymptotic stability.

For the bioinspired backstepping control, the time derivative of Lyapunov candidate function \eqref{22} is divided into three components, which would be easier for the calculations. These three components are, ${e_x}{{\dot e}_x} + {e_y}{{\dot e}_y}$, $\frac{k_a}{{B_1}}{L_1}{{\dot L}_1} + \frac{k_a}{{B_2}}{L_2}{{\dot L}_2}$, and ${e_\psi}{{\dot e}_\psi}+\frac{k_b}{{B_3}}{L_3}{{\dot L_3}}$. From the definition of the tracking errors, ${e_x}{{\dot e}_x} + {e_y}{{\dot e}_y}$, is rewritten as
\begin{equation}
    {e_x}{{\dot e}_x} + {e_y}{{\dot e}_y} = {e_x}({x_r} - {x_a}) + {e_y}({y_r} - {y_a})\label{23}
\end{equation}
Then, based on \eqref{eq2} and \eqref{jac}, \eqref{23} rewrites as,
\begin{equation}
\begin{gathered}
    \quad{e_x}{{\dot e}_x} + {e_y}{{\dot e}_y} =\\
    \quad{e_x}((C ({\psi _d}){u_d} - S ({\psi _d}){v_d}) - ({u_a}C ({\psi _a}) - {v_a}S ({\psi _a}))) + \\
    \quad{e_y}((S ({\psi _r}){u_r} + C ({\psi _r}){v_r}) - ({u_a}S ({\psi _a}) + {v_a}C( {\psi _a})))\label{24}
\end{gathered} 
\end{equation}
By substituting \eqref{11} into \eqref{24}, it can be calculated that 
\begin{equation}
\begin{split}
        {e_x}{{\dot e}_x} + {e_y}{{\dot e}_y} = {e_x}((C ({\psi _r}){u_r} - S ({\psi _r}){v_r}) - ({k_a}{L_1} + \\
        {u_r}C ({\psi _r}) + {v_r}S ({\psi _r}))) + {e_y}((S ({\psi _r}){u_r} + C ({\psi _r}){v_r}) -\\
        ({k_a}{L_2} + {u_r}S ({\psi _r}) - {v_r}C ({\psi _r}))) =  - {k_a}{e_x}{L_1} - {k_a}{e_y}{L_2} \label{18}
\end{split}
\end{equation}
As for the second part of the Lyapunov function for the bioinspired backstepping controller, $\frac{k_a}{{B_1}}{L_1}{{\dot L}_1} + \frac{k_a}{{B_2}}{L_2}{{\dot L}_2}$, by assuming $B_1=D_1$, $B_2=D_2$, based on the definition of the shunting model in \eqref{shut}, it is derived as
\begin{equation}
\begin{split}
\begin{array}{l}
\frac{{{k_a}}}{{{B_1}}}{{L_1}{{\dot L}_1} + \frac{{{k_a}}}{{{B_2}}}{L_2}{{\dot L}_2}}  = \\
\frac{{{k_a}}}{{{B_1}}}( - {A_1} - f({e_x}) - g({e_x}))L_1^2 + {k_a}(f({e_y}) - g({e_y})){L_1} + \\
\frac{{{k_a}}}{{{B_2}}}( - {A_1} - f({e_y}) - g({e_y}))L_2^2 + {k_a}(f({e_y}) - g({e_y})){L_2}\label{17}
\end{array}
\end{split}
\end{equation}
The last part of \eqref{22} is ${e_\psi}{{\dot e}_\psi}+{{k_b}}{L_3}{{\dot L_3}}$, which is calculated as
\begin{equation}
\begin{split}
        {e_\psi }{{\dot e}_\psi } + \frac{{{k_3}}}{{{B_1}}}{L_3}{{\dot L}_3} =  - {k_b}{e_\psi }{L_3}
        + \frac{{{k_b}}}{{{B_3}}}( - {A_3} - f({e_\psi }) \\- g({e_\psi }))L_3^2 + {k_b}(f({e_\psi }) - g({e_\psi })){L_3} \label{27}
\end{split}
\end{equation}
By combining \eqref{18}, \eqref{17}, and \eqref{27}, based on \eqref{shut} it can be concluded that $f(e_i)=e_i$ and $g(e_i)=0$,if ${e_i} \ge 0$. Otherwise if ${e_i} \le  0$ then $g(e_i)=e_i$ and $f(e_i)=0$. It can be concluded in \eqref{17} and \eqref{27} that
\begin{equation}
    \begin{array}{l}
-{A_i} - f({e_i}) - g({e_i}) \le 0\\
f({e_i}) - g({e_i}) - {e_i} = 0
\end{array}\label{20}
\end{equation}
In addition, as $t \to \infty$, $V_p \to 0$, overall, $\dot V_p$ is non-positive, if and only if $e_x$, $e_y$, and $e_\psi$ are zeros, $V_p=0$. Thus, the bioinspired backstepping control is proven to be asymptotically stable.

As for the bioinspired sliding mode control, substituting \eqref{shut} and \eqref{14} into \eqref{25}, it gets
\begin{equation}
\begin{gathered}
     {{\dot V}_z} = {S^T}(\ddot e + 2\Gamma \dot e + {\Gamma ^2}e) + \frac{1}{{{B_4}}}( - {A_4}^T{L_4}^T\\
      + ({B_4}^T - {L_4}^T)f{(S)^T} - ({D_4}^T + {L_4}^T)g{(S)^T}){L_4}\label{29}
\end{gathered}
\end{equation}
Then, substituting \eqref{1} and \eqref{16} into \eqref{29}, $\dot V_z$ becomes
\begin{equation}
\begin{gathered}
     {{\dot V}_z}{\rm{ = }}{S^T}{L_4} + \frac{1}{{{B_4}^T}}{L_4}^T( - {A_2}^T - \\
     f{({\rm{S}})^T} - g{(S)^T}) + \frac{1}{{{B_4}^T}}{L_4}({B_4}^Tf{(S)^T} - {D_4}^Tg{(S)^T}){L_4}\label{30}
\end{gathered}
\end{equation}
By letting ${B_4}^T={D_4}^T$, \eqref{30} finally becomes
\begin{equation}
\begin{gathered}
  {{\dot V}_z}{\rm{ = }}\frac{1}{{{B_4}^T}}{L_4}{L_4}^T( - {A_4}^T - f{({\rm{S}})^T} - g{(S)^T}) +
  \\(f{(S)^T} - g{(S)^T} + S^T){L_4}\label{31}   
\end{gathered}
\end{equation}
Similarly to the concept of \eqref{20}, it is obvious that $\frac{{L_4}{L_4}^T}{{{B_4}^T}}( - {A_4}^T - f{({S})^T} - g{(S)^T}) \le 0$ and $(f{(S)^T} - g{(S)^T} + S^T) = 0$. As $t \to \infty$, both $L_4$ and $S$ approaches zero, only if $e$ equals to zero, $S=0$. In addition, using the input-output property of shunting model from \eqref{shut} as $e \to 0$, $L_4$ approaches zero as well.
\begin{figure}[h!]
\centerline{\includegraphics[width=0.40\textwidth]{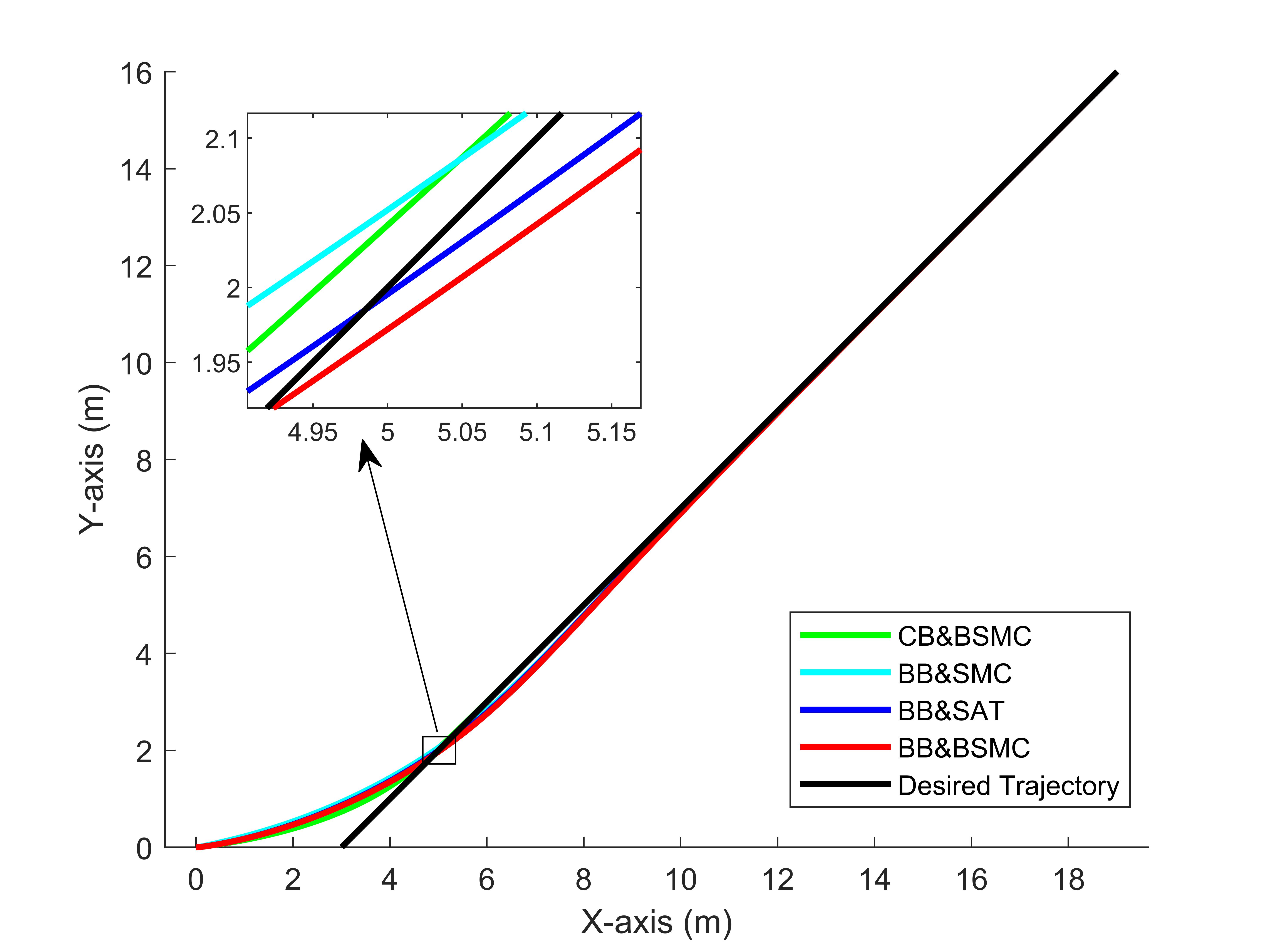}}
\caption{Straight path tracking trajectories. BB: Bioinspired backstepping, CB: Conventional backstepping, SMC: Conventional sliding mode, BSMC: Bioinspired sliding mode, SAT: Sliding mode with saturation \label{fig4}}
\end{figure}
For the overall stability, the Lyapunov candidate function and its time derivative are proposed as
\begin{equation}
\begin{split}
    V_o=V_p+V_z\\
    {{{\dot V}}_{o}} = {{{\dot V}}_{p}}{ + }{{{\dot V}}_{z}}
\end{split}
\end{equation}
Based on the results obtained from \eqref{23} to \eqref{31}, it is easy to verify that $V_o \le0$, if and only if $e_x$, $e_y$, $e_\psi$, and $e$ equal to zero, then $V_o=0$. Thus, the overall control stability has been proven. 
\section{Results}
This section demonstrates the superiority of the bioinspired aided controllers over the conventional method. The proposed control strategy generates smooth velocity and torque commands, which overcomes velocity overshoot and chattering problems in backstepping control and sliding mode control, respectively. Simulation results are shown into two different parts, tracking a straight path and a circular path, which are the two type of typical movements for a UUV operating underwater. In order to demonstrate the superiority of the proposed control strategy, three other different methods are shown in the comparisons. The UUV parameters \cite{Zhu2013TheVehicles,Sun2016AAccommodation} are provided as $m - {X_{\dot u}}=m - {Y_{\dot v}}=54.35$, ${I_z} - {N_{\dot r}} = 1.93$. The linear drag with respect to surge, sway, and yaw motion are treated as ${X_u} = {Y_v} = 17.51$, ${N_r} = 2.4$, finally the quadratic drag for the UUV are ${X_{uu}} = 10$, ${Y_{vv}} = 10$, and ${N_{rr}} = 2$ in surge, sway, and yaw motions, respectively.
\begin{figure}[h!]
\centerline{\includegraphics[width=0.40\textwidth]{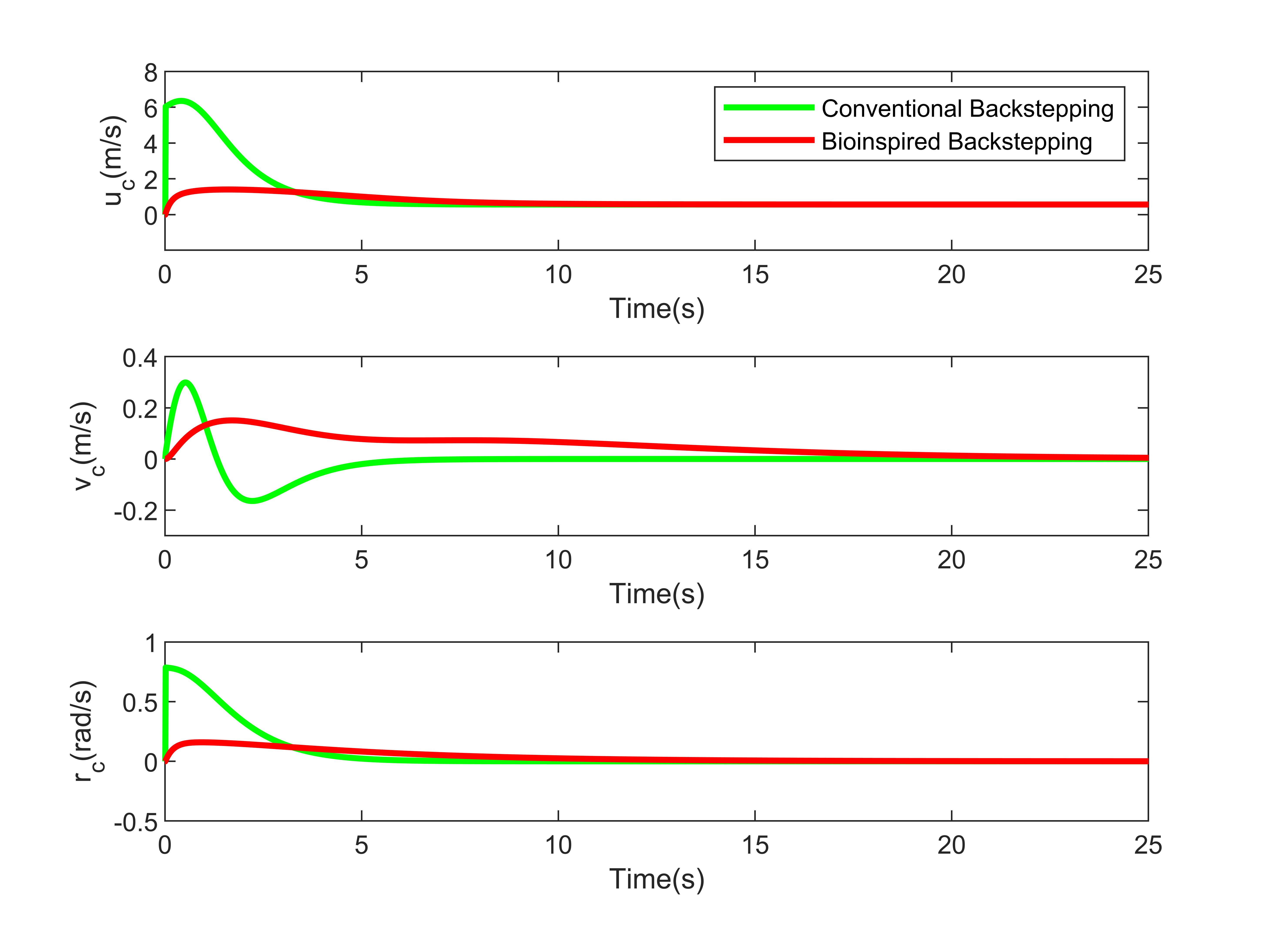}}
\caption{Velocity commands for UUV tracking a straight path. Green: Conventional backstepping; Red: Bioinspired backstepping  \label{fig5}}
\end{figure}
The control variables are set to $k=\left[{\begin{array}{*{20}{c}}2&2&0.2\end{array}} \right]^T$, $k_a=2$, $k_b=1$, $k_s=1$, $\Gamma=1$. As for the shunting model that is used in both backstepping and sliding mode control, the variables are set to $A_1=A_2=A_3=4$, $B_1=B_2=B_3=D_1=D_2=D_3=1$, ${A_4} = {\left[{\begin{array}{*{20}{c}}3&3&3\end{array}} \right]^T}$, $B_4=D_4={\left[{\begin{array}{*{20}{c}}1&1&1\end{array}} \right]^T}$. In addition, the saturation function that is used to replace the sign function in sliding is provided as ${{B}_{4}}{=}{{D}_{4}}{=}{\left[ {\begin{array}{*{20}{c}}1&1&0.03\end{array}} \right]^T}$, and $k_s=3$. The torque command follows the time response of the first order system to ensure its smoothness; then the actual torque commands generated are defined as ${\tau _c}(t) = \tau (t)(1 - {e^{{\raise0.7ex\hbox{${ - t}$} \!\mathord{\left/
 {\vphantom {{ - t} \sigma }}\right.\kern-\nulldelimiterspace}
\!\lower0.7ex\hbox{$\sigma $}}}})$, where $\tau_c$ is the actual command that is sent to the UUV dynamics and $\sigma$ is set to be 0.5. Finally, the sampling time is set to be 0.01 sec.
\begin{figure}[h!]
\centerline{\includegraphics[width=0.40\textwidth]{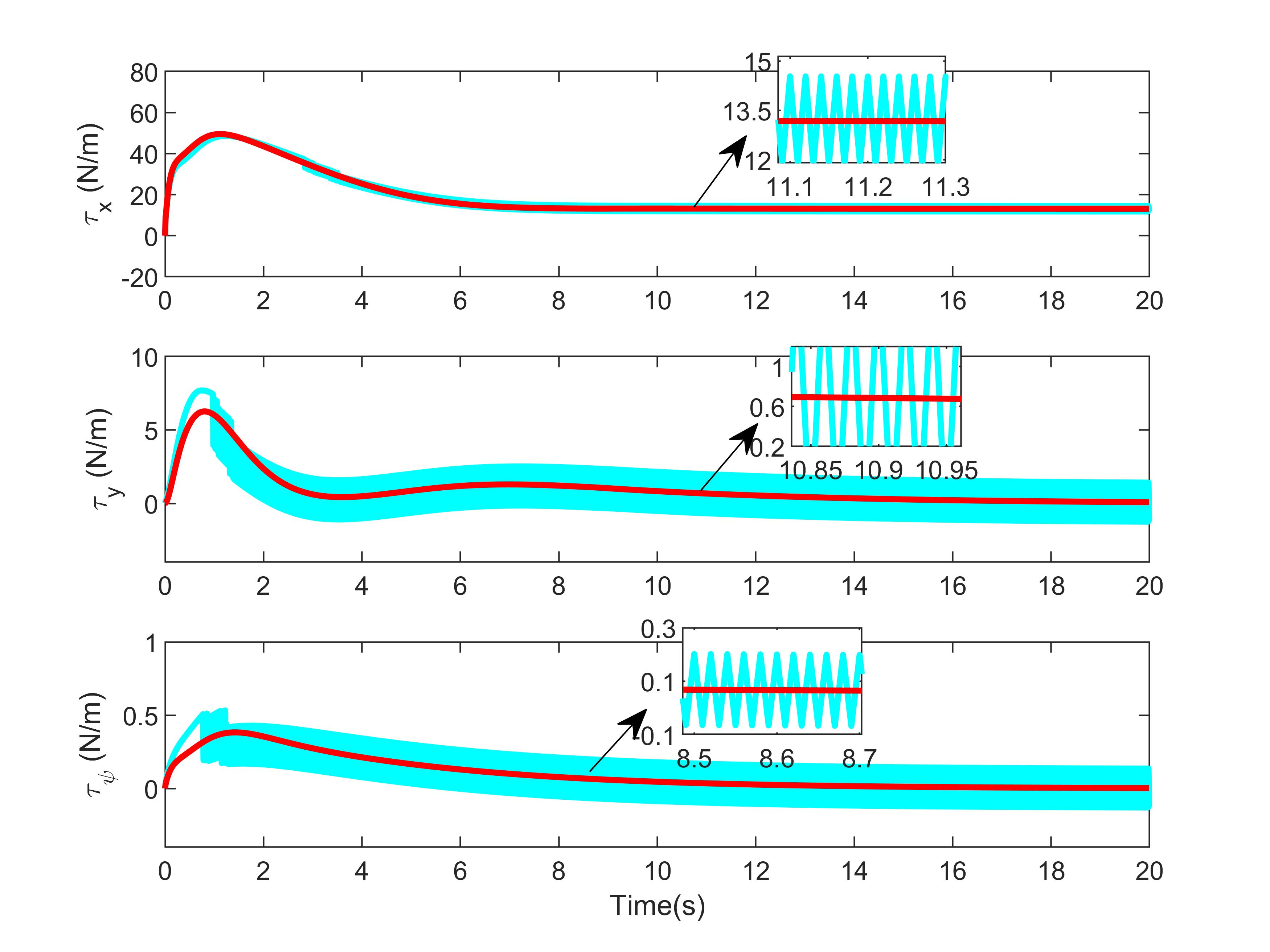}}
\caption{Torque commands for UUV tracking a straight path. Cyan: Conventional sliding mode control; Red: Bioinspired sliding mode control \label{fig6}}
\end{figure}

\subsection{Straight path tracking}
One of the most common movements for a UUV operating underwater is to move straight forward; therefore, a straight path is given to test the efficiency and effectiveness of the proposed method; in addition, it is assumed that the desired tracking trajectory is continuous and differentiable. Given the desired tracking trajectory as ${X_d} = 3 + 0.4t$, ${Y_d} = 0.4t$, ${\psi _d} = 45^\circ $, whereas the initial state for the UUV is set as $(0,0)$ and $\psi_a=0$. It is shown in Figure \ref{fig4} that all the control strategies track desired trajectory, and the conventional backstepping control strategy yields a slightly different path than other methods with bioinspired backstepping control.
\begin{figure}[h!]
\centerline{\includegraphics[width=0.4\textwidth]{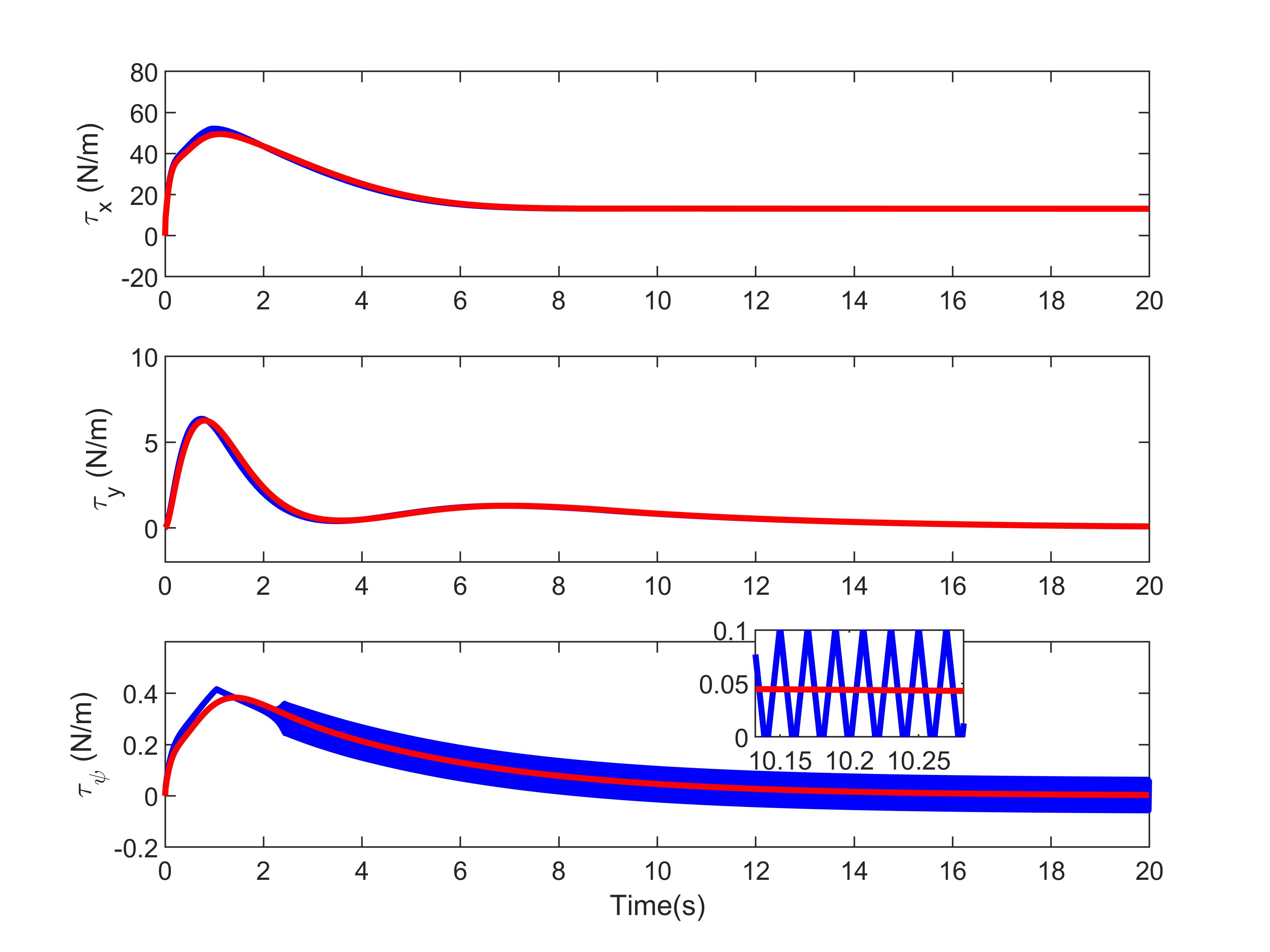}}
\caption{Torque commands comparison for UUV tracking a straight path. Blue: Sliding mode control with saturation function; Red: Bioinspired sliding mode control\label{fig7}}
\end{figure}
\begin{figure}[h!]
\centerline{\includegraphics[width=0.4\textwidth]{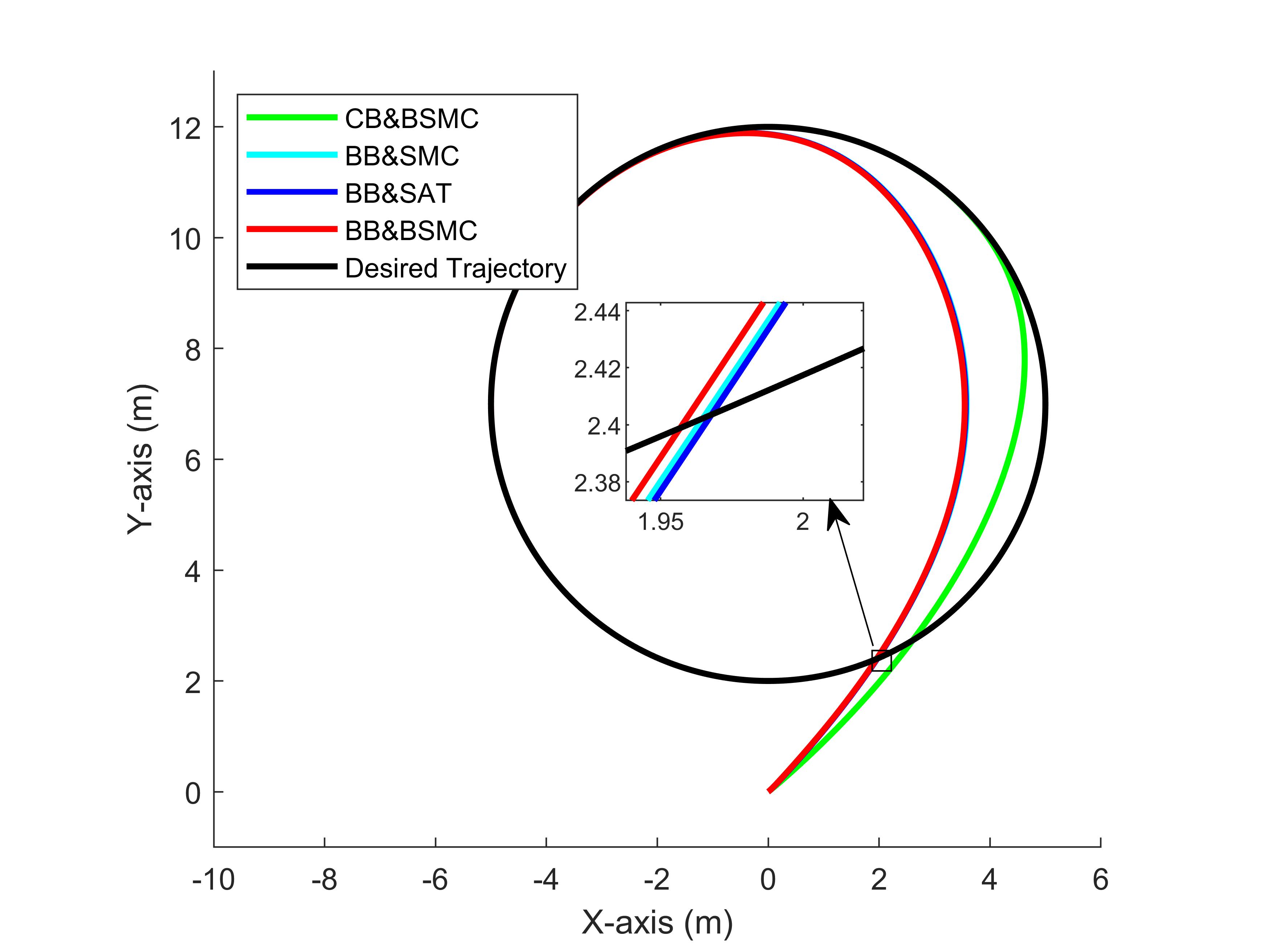}}
\caption{Circular trajectory tracking with different control strategies. BB: Bioinspired backstepping, CB: Conventional backstepping, SMC: Conventional sliding mode, BSMC: Bioinspired sliding mode, SAT: Sliding mode with saturation\label{fig8}}
\end{figure}
However, in Figure \ref{fig5}, the velocity commands that are generated from the conventional backstepping control suffers from the speed jump issues, whereas other control method that are implemented with bioinspired backstepping control yield smooth velocity commands without any speed jump. This speed jump issue is critical for UUVs in real world applications, because a speed jump would infer that the initial demanding torque would be infinitely large for a UUV to be able to reach such speed instantly. Therefore, the bioinspired backstepping control has practically solved the issue. In addition, based on the observation in Figure \ref{fig6}, the sliding mode torque control suffers from the chattering issue whereas the bioinspired sliding mode control has avoided chattering and provide smooth control inputs. 

\begin{figure}[h!]
\centerline{\includegraphics[width=0.4\textwidth]{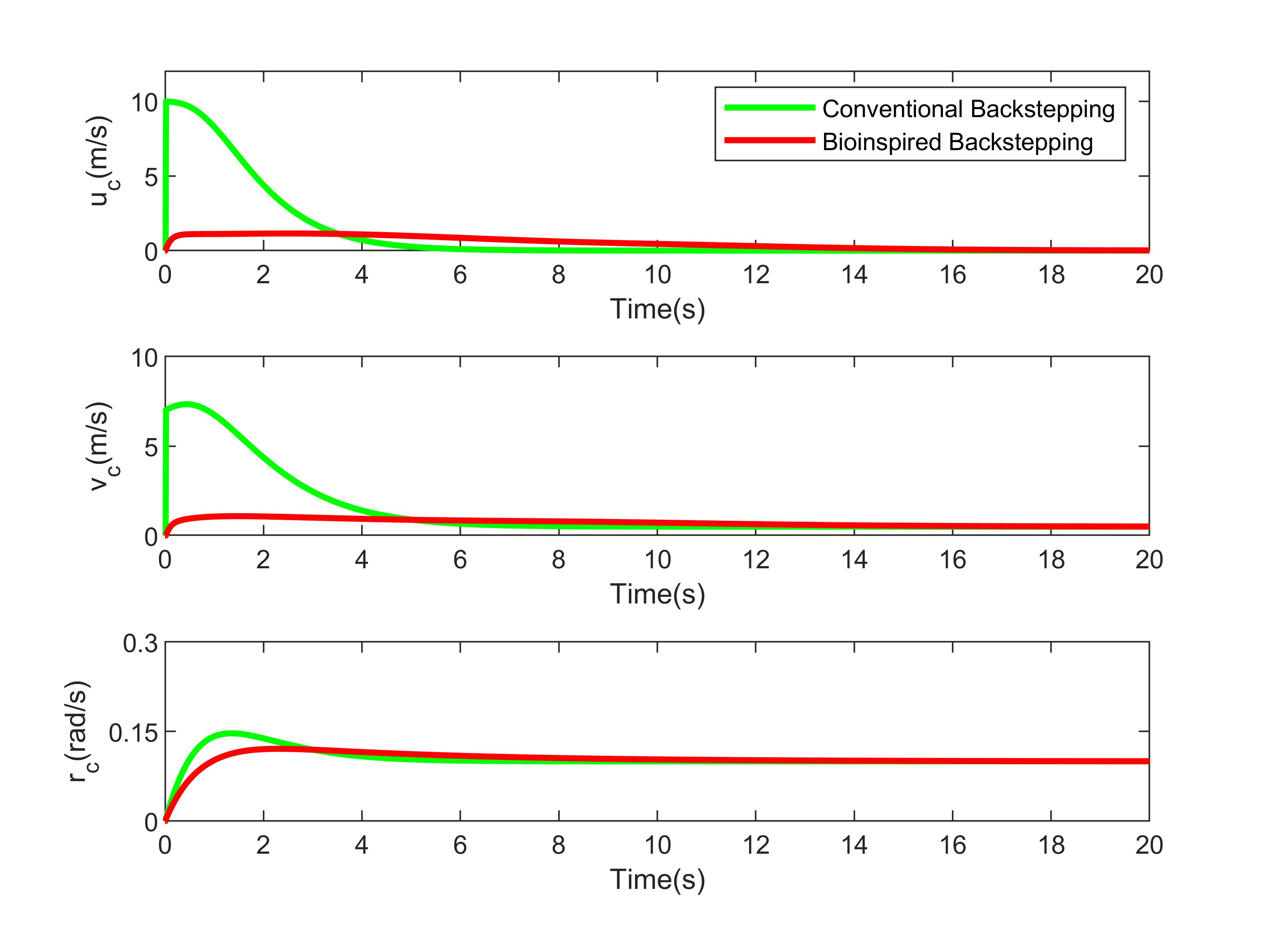}}
\caption{Torque commands for UUV tracking a circular path. Green: Conventional backstepping; Red: Bioinspired backstepping \label{fig9}}
\end{figure}
The saturation function is a commonly used method that can be integrated with sliding mode control to eliminate the chattering issue in conventional control method. Based on the definition of \eqref{sat}, the saturation function could still potentially produce chattering problems if the tracking error does not fall between $[-D_4,B_4]$ and the $k_s$ is not perfectly tuned. Since the UUV usually operates in complicated environments, in which the control parameters for sliding mode control with saturation function could be difficult to tune to give satisfactory results. As shown in Figure \ref{fig7}, although $\tau_x$ and $\tau_y$ are relatively smooth, there is still chattering issue which occurs in $\tau_\psi$. Therefore, the saturation function does not completely resolve the chattering issue but bioinspired sliding mode control would be able to handle the chattering problem perfectly.

\subsection{Circular path tracking}
This subsection tests the proposed tracking control strategy for a UUV tracking a circular path, which is a typical movement that UUVs need to achieve for trajectory tracking. The reference tracking path is defined as $X_r = 5\text {cos}(0.1t)$, $Y_r = 7+5\text{sin}(0.1t)$, and $\Psi_d=0.1t$, the starting state of the UUV is defined as $(0,0)$ and $\psi_a=0$. 
\begin{figure}[h!]
\centerline{\includegraphics[width=0.4\textwidth]{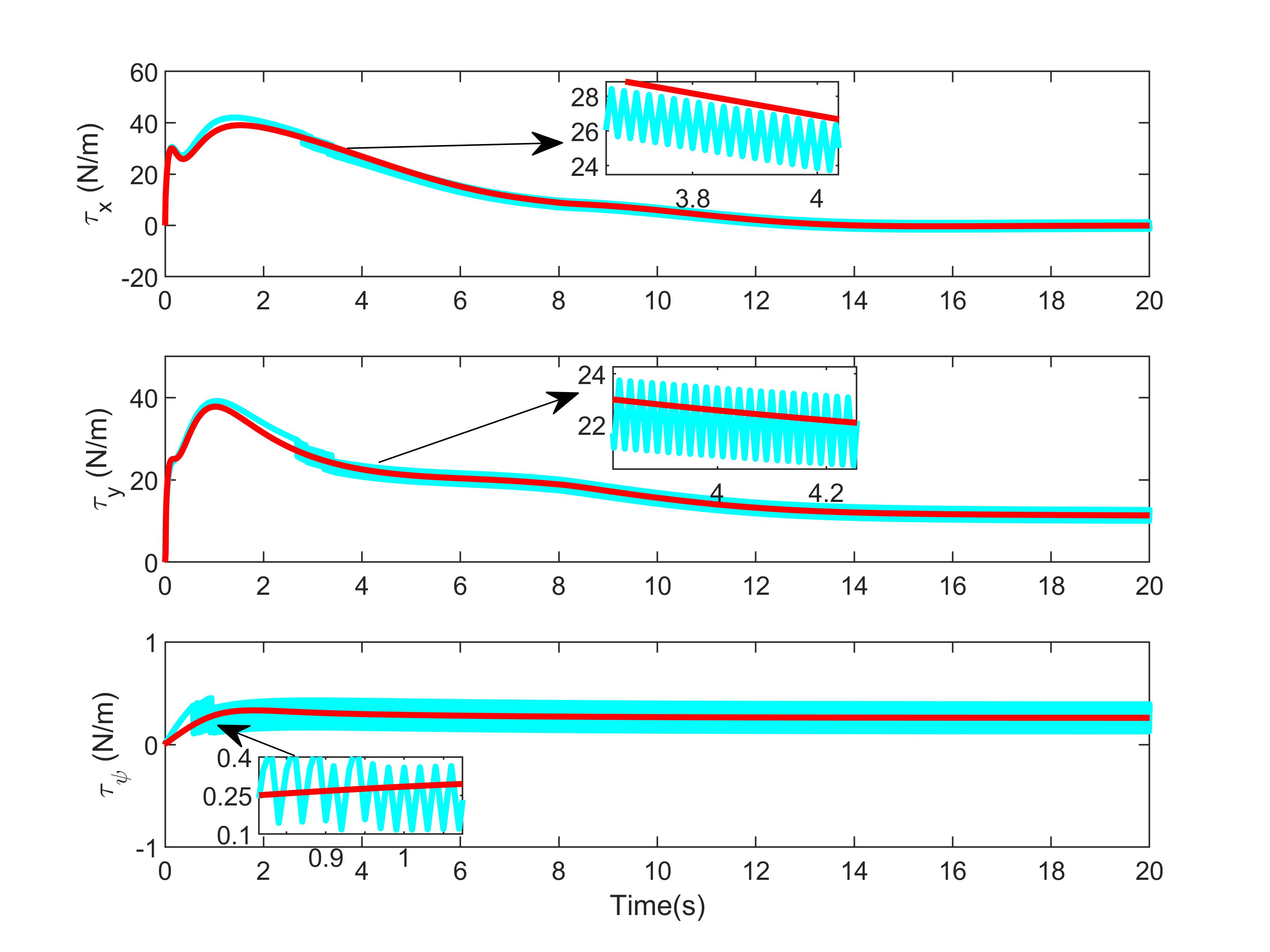}}
\caption{Torque Commands for Circular Path Tracking. Cyan: Conventional sliding mode control; Red: Bioinspired sliding mode control\label{fig10}}
\end{figure}
The results are shown in Figure \ref{fig8}, all the tracking strategies make the UUV tracks the desired circular path. However, from Figure \ref{fig9}, it once again shows that the bioinspired backstepping control outperforms the conventional backstepping control, the bioinspired backstepping control once again provides smooth velocity commands without sharp velocity changes whereas there is large velocity jumps from the conventional method.
\begin{figure}[h!]
\centerline{\includegraphics[width=0.4\textwidth]{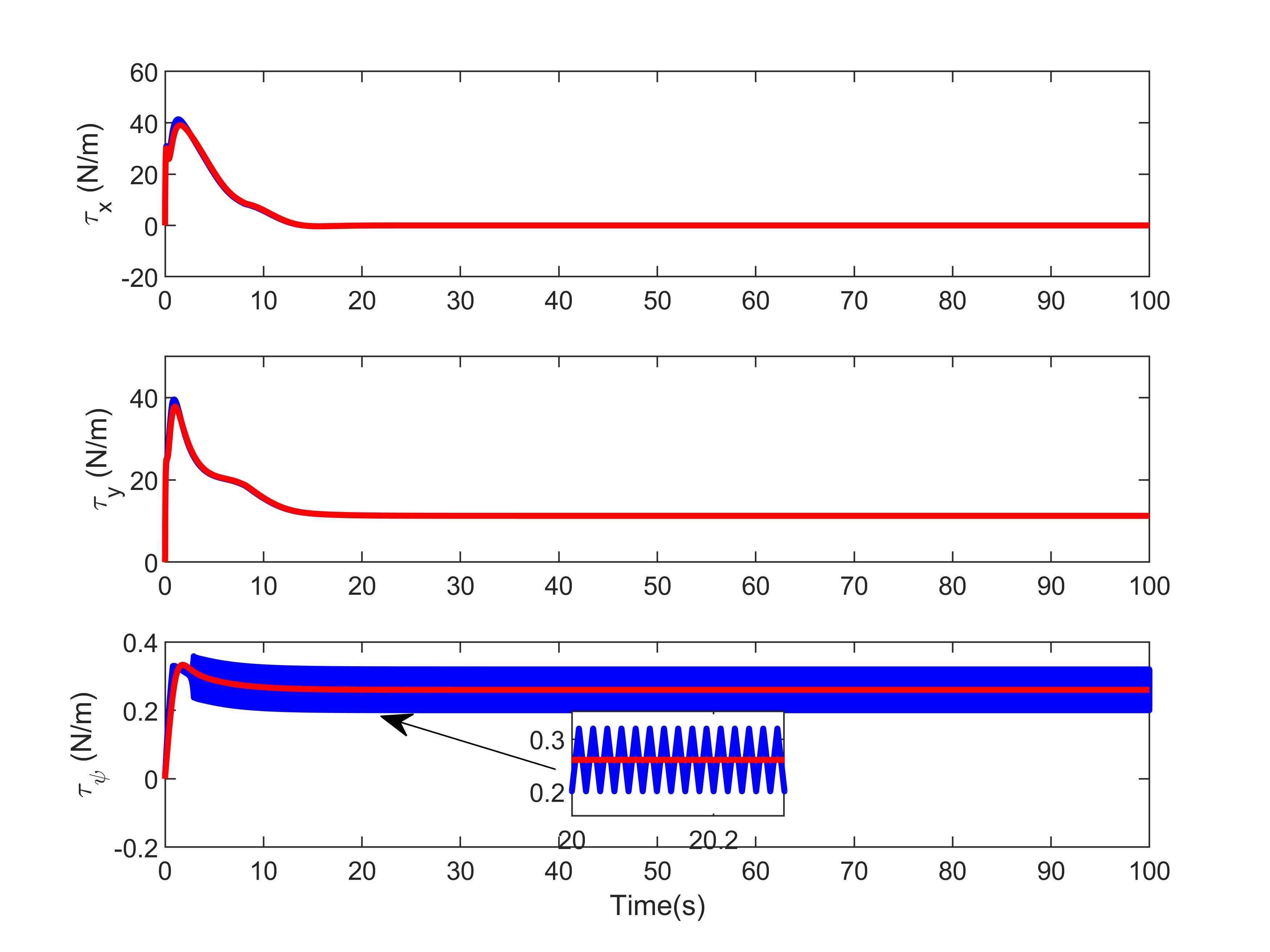}}
\caption{Torque commands comparison for UUV tracking a circular path Blue: Sliding mode control with saturation function; Red: Bioinspired sliding mode control\label{fig11}}
\end{figure}

The velocity commands from the bioinspired backstepping control method then propagate through the dynamic controller, where the results are shown in Figure \ref{fig10}. The output from the sliding mode controller suffers chattering issue and bioinspired sliding mode controller yield smooth torque commands. In addition, by comparing the results with the saturation function in Figure \ref{fig11}, it shows that tracking a circular path for a UUV does not remove the chattering issue in the saturation function, while bioinspired sliding mode control still yields satisfactory results. The proposed hybrid control strategy has proven to be the best choice over the other control strategies in the provided results.
\subsection{Tracking performance under noises}
The UUV often operates in complex and complicated environments where the system and measurement noises play major roles in tracking performance, as well as the control smoothness. Thus, the proposed control strategy is tested under system and measurement noises. The Kalman filter and extended Kalman filter are respectively used to provide accurate state estimates for the kinematics and dynamics of the UUV. The system noises are considered as zero mean Gaussian with the covariance of $[\begin{array}{*{20}{c}}
{{{10}^{ - 3}}}&{{{10}^{ - 3}}}&{{{10}^{ - 4}}}
\end{array}]$ in the velocities estimates, and $[\begin{array}{*{20}{c}}
{{{10}^{ - 5}}}&{{{10}^{ - 5}}}&{{{10}^{ - 6}}}
\end{array}]$ in the position state estimates. The measurement noises are treated as ten times higher than the system noises, respectively. The path that is used to test the tracking performance is a circular path same as Subsection 4.2, which is defined as $X_r = 5\text {cos}(0.1t)$, $Y_r = 7+5\text{sin}(0.1t)$, and $\Psi_d=0.1t$, the starting state of the UUV is defined as $(0,0)$ and $\psi_a=0$.

\begin{figure}[h!]
\centerline{\includegraphics[width=0.4\textwidth]{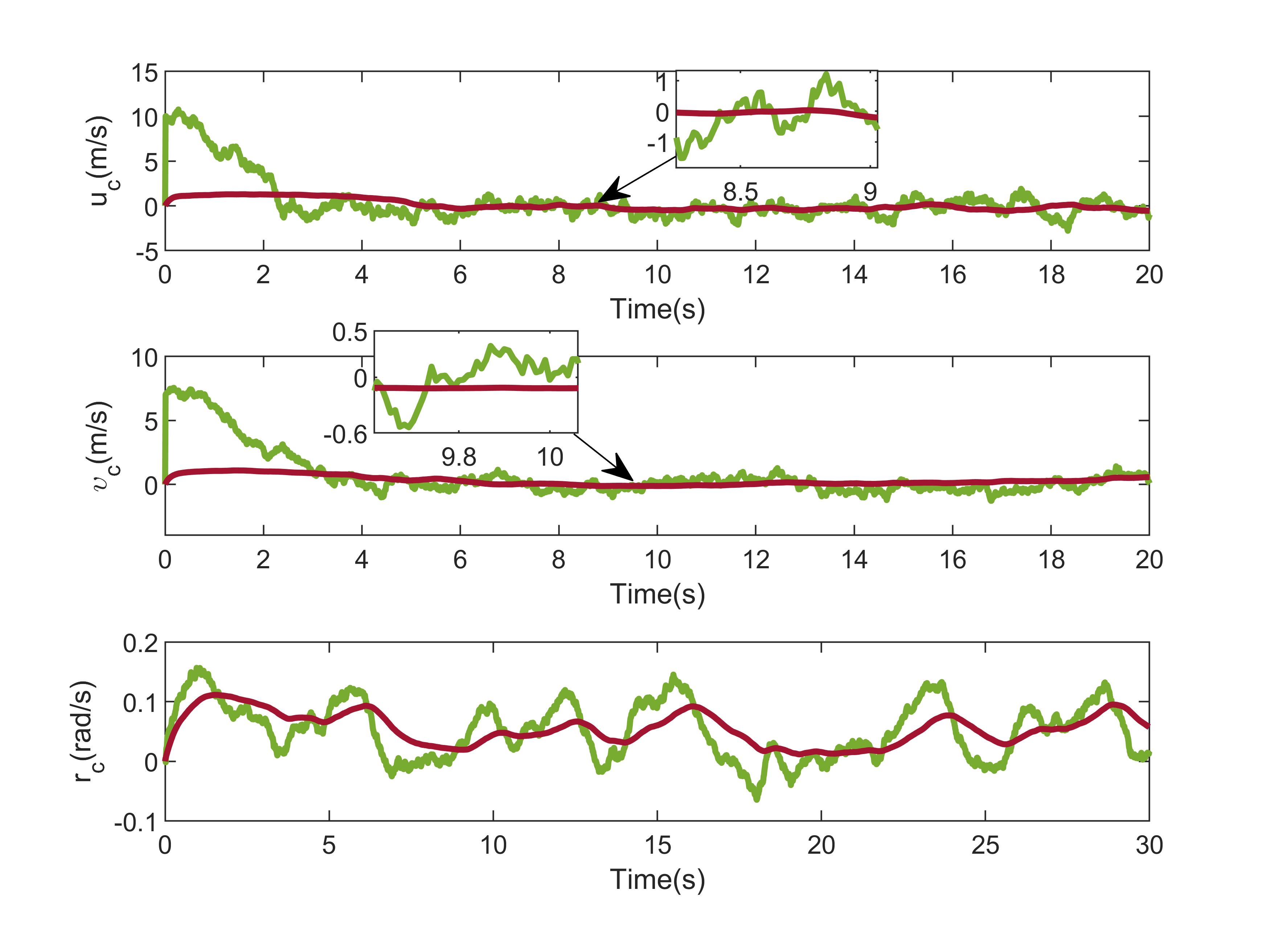}}
\caption{Velocity commands comparison for UUV tracking a circular path under noises Green: Conventional Backstepping and Sliding mode control with saturation function; Red: Bioinspired Backstepping and Bioinspired sliding mode control\label{fig12}}
\end{figure}

As shown in Figure \ref{fig12}, under the noises, the proposed bioinspired control strategy is capable of providing smooth velocity command while the control signal in conventional method has chattering issue under the noises. This control smoothness is critical since the actuator may not be able to handle fast changing signal, and the conventional method apparently gives extra burden to the actuator.
\section{Conclusion}
This paper developed a hybrid control strategy for UUV, the proposed control strategy contains a bioinspired backstepping kinematic controller and bioinspired sliding mode controller, which respectively resolves speed jump and chattering issues in conventional design, respectively. In addition, the proposed strategy is capable of providing smooth control inputs even under system and measurement noises. To demonstrate the efficiency and effectiveness of the proposed method, a comprehensive comparison study is conducted, the results show that the proposed control strategy has more advantages over the other method in terms of the tracking command smoothness and practicability. Further studies could potentially focus on resolving the effects of disturbances with accurate state estimates toward the proposed control strategy.

\section{Acknowledgments}

This work is supported by the Advanced Robotic Intelligent Systems Laboratory at the University of Guelph under Natural Sciences and
Engineering Research Council of Canada (NSERC)

\bibliographystyle{IEEEtran}
\bibliography{IEEEabrv,references}

\end{document}